\documentclass[journal]{IEEEtran}
\usepackage{graphicx}
\usepackage[numbers,sort&compress]{natbib}
\usepackage{algorithm,algpseudocode,algorithmicx}
\usepackage{amsmath,latexsym,amssymb,amsthm,array,amsfonts,algorithm,algpseudocode,booktabs,graphicx,subfigure,multirow,cuted,stfloats}
\ifCLASSINFOpdf
\else
\fi
\begin{document}
%
\title{Sampled Training and Node Inheritance for Fast Evolutionary Neural Architecture Search}
%
%
%

\author{Haoyu~Zhang,
        Yaochu~Jin,~\IEEEmembership{Fellow,~IEEE,}
        Ran~Cheng,~\IEEEmembership{Member,~IEEE,}
        and~Kuangrong Hao
\thanks{H. Zhang, Y. Jin and K. Hao are with the Engineering Research Center of Digitized Textile \& Apparel Technology, Ministry of Education, College of Information Science and Technology, Donghua University, Shanghai 201620, China. Email: zhy920816@sina.cn; krhao@dhu.edu.cn}
\thanks{Y. Jin is with the Department of Computer Science, University of Surrey, Guildford, Surrey GU2 7XH, United Kingdom. He is also with the Engineering Research Center of Digitized Textile \& Apparel Technology, Ministry of Education, College of Information Science and Technology, Donghua University, Shanghai 201620, China. Email: yaochu.jin@surrey.ac.uk. (\textit{Corresponding author})}
\thanks{R. Cheng is with the Shenzhen Key Laboratory of Computational Intelligence, University Key Laboratory of Evolving Intelligent Systems of Guangdong Province, Department of Computer Science and Engineering, Southern University of Science and Technology, Shenzhen 518055, China. Email: chengr@sustech.edu.cn.}
\thanks{Manuscript received February 14, 2020; revised xxxx, 2020.}}

%
%

\markboth{Journal of \LaTeX\ Class Files,~Vol.~xx, No.~x, August~20xx}%
{Shell \MakeLowercase{\textit{et al.}}: Bare Demo of IEEEtran.cls for IEEE Journals}
%



\maketitle

\begin{abstract}

The performance of a deep neural network is heavily dependent on its architecture and various neural architecture search strategies have been developed for automated network architecture design. Recently, evolutionary neural architecture search (ENAS) has received increasing attention due to the attractive global optimization capability of evolutionary algorithms. However, ENAS suffers from extremely high computation costs because a large number of performance evaluations is usually required in evolutionary optimization and training deep neural networks is itself computationally very intensive. To address this issue, this paper proposes a new evolutionary framework for fast ENAS based on directed acyclic graph, in which parents are randomly sampled and trained on each mini-batch of training data. In addition, a node inheritance strategy is adopted to generate offspring individuals and their fitness is directly evaluated without training. To enhance the feature processing capability of the evolved neural networks, we also encode a channel attention mechanism in the search space. We evaluate the proposed algorithm on the widely used datasets, in comparison with 26 state-of-the-art peer algorithms. Our experimental results show the proposed algorithm is not only computationally much more efficiently, but also highly competitive in learning performance.

\end{abstract}


%
\IEEEpeerreviewmaketitle

\section{Introduction}
%
%
%
%

\IEEEPARstart{D}{EEP} learning has achieved remarkable success in solving various tasks such as image classification \cite{sainath2013deep}, speech recognition \cite{abdelhamid2014convolutional}, natural language processing  \cite{sutskever2014sequence}, among many others. Since the performance of deep neural networks heavily depends on their architecture, a large body of research effort in the deep learning community has been dedicated to the design of novel architectures such as DenseNet \cite{huang2017densely}, ResNet \cite{he2016deep, he2016identity}, and VGG \cite{simonyan2014very}. In general, most powerful deep networks were manually designed by human experts who have extensive expertise in both deep learning and the related problem domain. Not until recently has automated neural architecture design, i.e., neural architecture search (NAS), shown great opportunity to allow interested users without adequate domain knowledge to benefit from the success of deep neural networks \cite{zoph2016neural}.

An NAS task can generally be formulated as a complex optimization problem \cite{elsken2018neural, khan2019survey}. In the field of computational intelligence, evolutionary algorithms (EAs) \cite{back1996evolutionary} have widely been used to solve various neural network training problems \cite{yao1999evolving}, such as weight training \cite{whitley1990genetic}, architecture design \cite{xie2017genetic}, and learning rule adaptation \cite{wang2019evolving}. Most recently, evolutionary neural architecture search (ENAS) employing an EA as the optimizer for NAS received increasing attention \cite{real2017large-scale, suganuma2017genetic, sun2019surrogate, sun2018automatically}. Despite that EAs have shown strong search performance on a variety of optimization tasks \cite{deb2002fast, wang2014two_arch2, sun2018improved, sun2019igd}, they generally suffer from high computation costs as a class of population-based search methods. This is particular true for ENAS since EAs typically require a large number of fitness evaluations, and each fitness evaluation in NAS is computationally intentive as it usually involves the training of a deep neural network from scratch on a large amount of data. For example, it takes 22 GPU days with three 1080TI GPUs for AE-CNN \cite{sun2018automatically} to obtain an optimized CNN architecture on CIFAR10 dataset.

Therefore, various techniques have been suggested in ENAS to reduce the computation costs without seriously degrading the optimization performance. For example, low fidelity estimates of the performance are commonly used, which unfortunately substantially deteriorate the search performance \cite{elsken2018neural, zela2018towards}.  In \cite{swersky2014freeze}, Bayesian optimization \cite{BO} is used to speed up evolutionary optimization, which is called Freeze-thaw Bayesian optimization. The main idea is to build a model to predict the performance based on the training performance in the previous epochs. Unfortunately, this algorithm is based on Markov chain Monte Carlo sampling and also suffers from high computational complexity. Recently, Sun et al. proposed a surrogate-assisted ENAS termed E2EPP, which is based on a class of surrogate-assisted evolutionary optimization \cite{sun2019surrogate, jin2019data-driven, jin2011surrogate, wang2018offline} that was meant for data-driven evolutionary optimization of expensive engineering problems. Specifically, E2EPP builds a surrogate that can predict the performance of a candidate CNN, thereby avoiding the training of a large number of neural network during the ENAS. Compared with AE-CNN, a variant of AE-CNN assisted by E2EPP (called AE-CNN+E2EPP) can reduce 214\% and 230\% GPU days on CIFAR10 and CIFAR100, respectively. However, AE-CNN+E2EPP still requires 3 GPUS for 17 days to achieve its best results since training sufficiently accurate surrogates can still be computationally expensive since it requires to train a large number of deep neural networks.

In addition to surrogate-assisted ENAS, several ideas of information reuse have been proposed to reduce computation costs, including parameter sharing \cite{pham2018efficient} that forces all sub-models to share a set of weights, knowledge inheritance \cite{zhang2018finding} that makes the child model directly inherit the weight of the convolution kernel of the parent model \cite{zhang2018finding}, and informed mutation \cite{real2017large-scale} that is designed to facilitate weight sharing. Another related work is known as network morphism \cite{jin2018efficient}, which aims to keep the functionality of a neural network while changing its architecture.

Apart from automated neural architecture design using ENAS, other methodologies such as attention mechanisms \cite{fu2017look} have been investigated to improve the performance of deep neural networks such as convolutional neural networks (CNNs). Inspired by the human visual system, attention mechanisms attempt to recognize objects by selecting the key parts of an object instead of the whole object \cite{zhao2017diversified, rensink2000dynamic}. Consequently, introducing an attention mechanism into CNNs can bring more discriminative feature representation capability \cite{wang2017residual}. In general, attention mechanism based methods can be divided into two categories. The first category includes methods focusing on channel attention, such as SE-Net \cite{hu2018squeeze}, ECA-Net \cite{wang2019eca-net:}, while the second category refers to methods based on spatial attention, such as spatial transformer networks \cite{jaderberg2015spatial}, and deep recurrent attentive writer (DRAW) neural network \cite{gregor2015draw}. Compared with spatial attention, channel attention can be more easily incorporated into CNNs for them to be trained end-to-end. Specifically, a channel attention module contains at least two branches: a mask branch and a trunk branch. The trunk branch performs feature transmitting or processing, while the mask trunk generates and learns weights of the output channels. The principle of the channel attention mechanism is to reconstruct channel-wise features, i.e., assigning a new weight to each channel to make the feature response of the key channel stronger, so that it can learn to selectively emphasize significant informative features and penalize excessive redundant features \cite{hu2018squeeze}. However, most previous work merely constructed attention mechanisms by manually stacking multiple attention modules into neural networks. Such naive stacking attention modules may lead to poor performance of the network \cite{wang2017residual}.

To improve the efficiency of ENAS, this paper proposes a fast ENAS framework based on sampled training of the parent individuals and node inheritance for generating offspring individuals, called SI-ENAS, thereby significantly reducing the computation costs. The main contributions of this paper are summarized below:
\begin{itemize}
\item A sampling technique is proposed to train the individual networks in the parent population. Specifically, for each mini-batch of the training data, a parent individual is randomly chosen and trained on the mini-batch. Since the  batch size of the training data is usually much larger than the population size, each individual in the parent population will be trained sequentially on a number of randomly chosen mini-batches of the training data.
\item A node inheritance strategy is proposed to generate offspring individuals by applying a one-point crossover and an exchange mutation. This way, offspring individuals directly inherit the parameters from their parents and none of them needs to be trained from scratch for evaluating their fitness value.
\item A multi-scale channel attention mechanism is incorporated into neural architecture search. As a result, channel-wise features can be presented with respect to the spatial information on different scales.
\item We empirically demonstrate that the proposed SI-ENAS method cannot only achieve excellent performance on CIFAR-10 and CIFAR-100, but also significantly reduces the computation cost of the architecture search process. Moreover, we show that the neural architecture designed for CIFAR-10 by SI-ENAS can be transferred to more a challenging classification task and achieve highly competitive results.
\end{itemize}

The rest of this paper is organized as follows. Section II introduces the background of this work. Section III describes the DAG-based neural network architecture encoding and the channel attention mechanism, followed by the details of the proposed algorithm in Section IV. Experimental settings and experimental results are presented in Section V and Section VI, respectively. Finally, Section VII concludes the paper.

\section{Preliminaries}

In this section, we briefly review the basic background of convolutional neural networks, channel attention mechanisms, and evolutionary optimization.

\subsection{Convolutional neural networks}

Convolution extracts locally correlated features by dividing the image into small slices, making it capable of learning suitable features \cite{khan2019survey}. The convolutional layer uses the convolution kernel recognized as an array of square block neurons to implement convolutional operations on the input data. For example, given an input $ x=[x_1,x_2,\cdots,x_c]$, $k=[k_1,k_2, \cdots ,k_s]$ is used to denote the set of convolution kernels, where $s$ refers to the number of filter. Let $\sigma$ be the activation function and '*' be a convolution operation, then the output $ \hat{y}=[\hat{y}_1,\hat{y}_2,\cdots,\hat{y}_s]$ is computed by a nonlinear activation function as follows:
\begin{align}\label{1}
  \hat{y} = \sigma(k*x)\,\, ,
\end{align}

The pooling layer is a non-linear down-sampling operation, which can be added to CNNs after several convolutional layers. Generally, there are two types of pooling layers in CNNs, max pooling and average pooling. Hence, the output of the filter is the maximum or mean value of the area. Subsequently, the feed-forward propagation passes through several convolution and pooling layers and outputs the result of classification. Usually, the rectified linear units (ReLU) \cite{hochreiter1998vanishing} is used as the activation function in the fully connected classification layers and the softmax function is employed as the activation function in the output layer:
\begin{align}\label{2}
  \sigma_{ReLU}(h)&=\max(0,h)\\
  \sigma_{softmax}(h_j)&=\frac{\exp(h_j)}{\sum_{i=1}^{n}exp(h_j)} \,\, ,
\end{align}
where $h$ is the output of the previous layers and $n$ is the total number of class labels. In image classification, the cross-entropy is commonly utilized as the loss function $\mathcal{L}(w)$ to be minimized:
\begin{align}\label{3}
\mathcal{L}(w)&=-\sum_{\varphi}y \log \hat{y} \\
\min \mathcal{L}(w)&=\frac{1}{M}\sum_{\l}\mathcal{L}(w,x_l)\quad x_l \in \{x_1,x_2,\cdots,x_M\} \, ,
\end{align}
where $w$ are the trainable parameters, i.e., weights and bias, $x_l$ is the $l$-th training sample, and $M$ is the size of training data.

The mini-batch stochastic gradient descent (mini-batch SGD) is adopted in this work, which randomly chooses the mini-batch size of the training data for computing the gradient. This approach aims to balance the computation efficiency and training stability in each training iteration:
\begin{align}\label{4}
g_i&=\frac{1}{b}\nabla_{w}(w,x_{i:i+b})\\
{w}_{i+1}&={w}_i-\eta*g_i \, \, ,
\end{align}
where $\eta$ is the learning rate, $b$ is the size of mini-batch, and $g_i$ is the average gradient over data samples $x_{i:i+b}$ with respect to the elements in $w_i$ in the $i$-th iteration. Parameter $w$ will be updated by iteratively subtracting $\eta*g_i$ from the current model parameter in training the neural network.

\subsection{Channel attention}

Incorporating channel attention mechanisms into CNNs has been shown very promising for performance improvement \cite{wang2019eca-net:, hu2018gather, hu2018squeeze, huang2019ccnet}. Among these mechanisms, the squeeze-and-excitation network (SENet) \cite{hu2018squeeze} is one of the competitive structures, which learns channel attention for the convolution layer. Despite their promising capability in performance enhancment, these methods are computationally intensive \cite{woo2018cbam, fu2019dual}. To address this problem, Wang et al. \cite{wang2019eca-net:} proposed an efficient channel attention (ECA) module that involves a small number of parameters by using a fast $1D$ convolution layer to capture cross-channel interactions. In addition, Wang et al. \cite{wang2017residual} introduced the idea of attention residual learning to improve the performance of attention mechanisms. Although the attention module works as feature selectors that enhance good channels or features, it can potentially damage the useful properties of original feature maps. Hence, attention residual learning uses residual connections \cite{he2016deep} to pass the original features forward to the deeper layers, which can enhance feature selection while keeping good properties of the original features.

\subsection{Evolutionary Neural Architecture Search (ENAS)}
\begin{figure}[H]
\centering 
\includegraphics[width=0.5\textwidth]{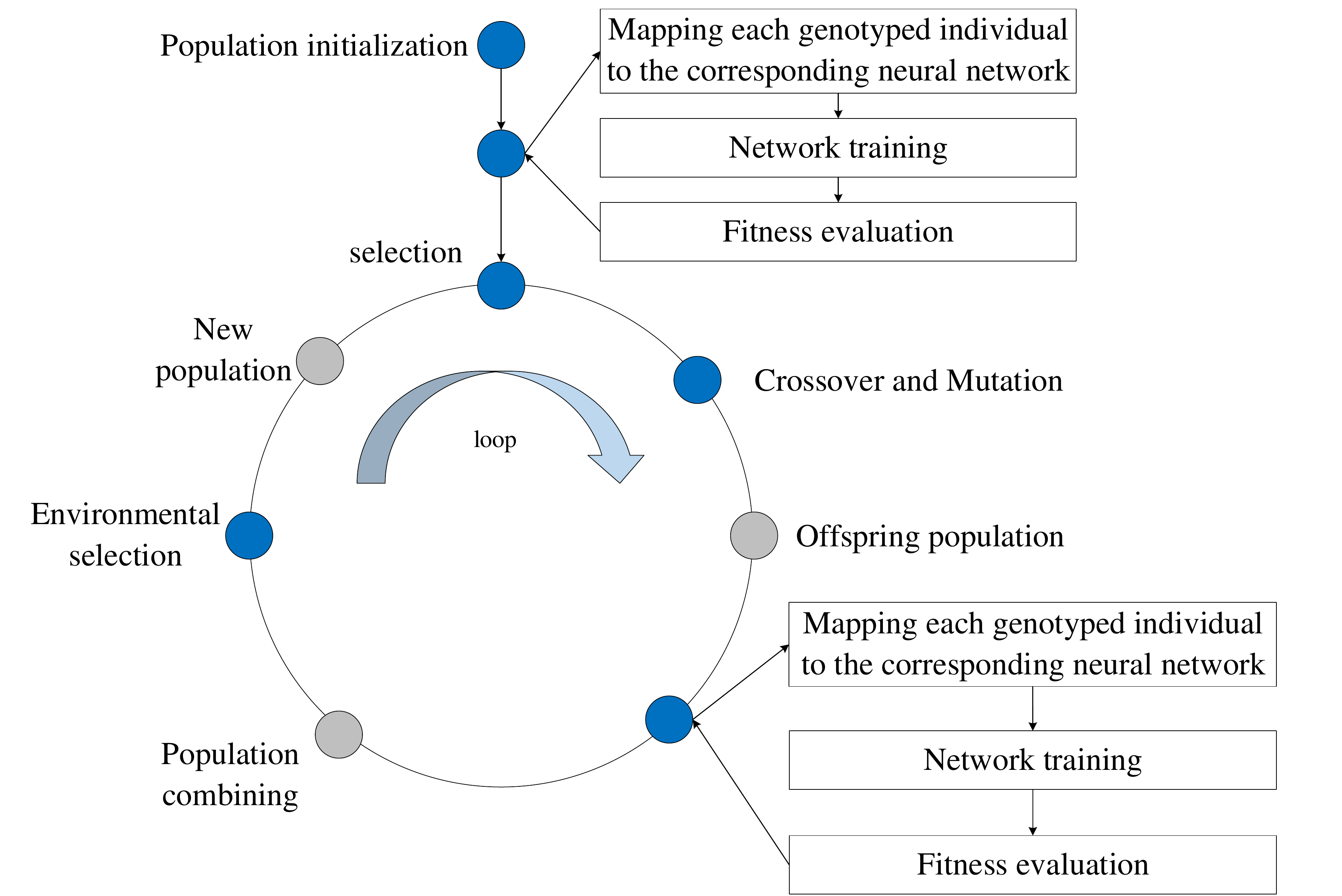} 
\caption{A generic ENAS framework.} 
\label{Fig_1}
\end{figure}

Fig.~\ref{Fig_1} shows a generic framework of ENAS \cite{sun2018automatically, sun2019surrogate, xie2017genetic}, which consists of the following six main steps:

\begin{enumerate}
  \item Randomly generate $K$ neural networks for the initial population $P_0$ based the corresponding network encoding strategy.
  \item Evaluate the fitness of each individual (neural network) in $P_t$ by training the network on a set of given data. The fitness function is usually a loss function to be minimized.
  \item Generate offspring (new candidate neural networks) from parent individuals using genetic operators such as crossover and mutation. Offspring population $Q_t$ has the same size as the parent population $P_t$.
  \item Evaluate the fitness of the generated offspring $Q_t$ and merge it with parent population $P_t$ into a combined population $R_t$, i.e., $R_t=P_t+Q_t$ has a size of $2K$. Note that in ENAS, the parent individuals sometimes also need to be trained and evaluated before environmental selection to avoid bias towards the offspring individuals.
  \item The parent population for the next generation $P_{t+1}$ is obtained by selecting $K$ better solutions from $R_t$ using an environmental selection method.
  \item Go to Step 3 if the evolution is not terminated; otherwise, select the best individual (neural network) in the parent population as the final solution.
\end{enumerate}

From the above, we can see that in general, ENAS follows the basic steps of an evolutionary algorithm (EA) \cite{back1996evolutionary, banzhaf1998genetic}, i.e., population initialization, reproduction, fitness evaluation, and environmental selection \cite{schmitt2001theory}. For fitness evaluations in ENAS, a neural network is trained on a training dataset and then evaluated a validation dataset to avoid overfitting. Hence, fitness evaluations in ENAS may take hours if the network is large and if the training dataset is huge. Since EAs are a type of population-based search method, they often require a large number of fitness evaluations, making ENAS computationally very expensive. For instance, on the CIFAR10 and CIFAR100 datasets, CNN-GA \cite{sun2018automatically} consumed 35 days and 40 days on 3 GPUs, respectively, the genetic CNN \cite{xie2017genetic} spent 10 days on 10 GPUs, and the large-scale evolutionary algorithm \cite{real2017large-scale} consumed 22 days on 250 GPUs. Therefore, it is essential to accelerate fitness evaluations in ENAS when computational resource is limited.

\section{Architecture Search Space}

\begin{figure*}
\centering 
\includegraphics[height=3.7cm, width=1\textwidth]{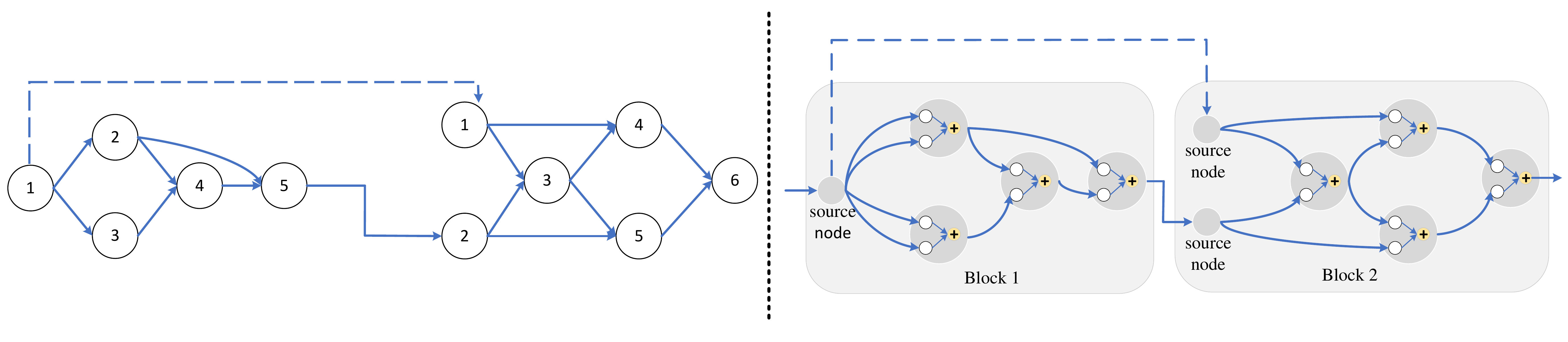} 
\caption{An example of two blocks in the defined search space with four computational nodes. Left: A computational DAG topology composed of two blocks. Right: the corresponding neural architecture consisting of two blocks. The first block contains one source node (node\,1) and four computational nodes, node\,5 is the block's output. The second block contains two source nodes (node\,1 and node\,2) and four computational nodes, node\,6 is the block's output. Here, the solid arrows represent the information flow in the DAG and the dotted arrows represent skip connections.} 
\label{Fig_2}
\end{figure*}

In this section, we describe the neural architecture encoding method used in this work, which defines the architecture search space. Our encoding method is built upon the micro search spaces proposed in \cite{pham2018efficient}, which is represented using a single directed acyclic graph (DAG). We first design smaller convolutional modules, denominated blocks, and then stack them together to create the overall neural network.

\subsection{Block structure}

A block is a fully convolutional network. Fig. \ref{Fig_2} shows an illustrative example of a computational DAG topology consisting of two blocks (left panel), and the corresponding neural architecture (right panel). Each block consists of source nodes and $N_c$ computation nodes. The source node is treated as the block's input, which is the output of the previous blocks or the input of the overall network. Each computation node consists of two computational operations and ends with an element-wise addition operation. Overall, we can describe a node $d$ in a block $i$ in the search space with a 5-tuple, ($\mathcal{I}_1,\mathcal{I}_2,\mathcal{O}_1,\mathcal{O}_2,\mathcal{M}$), where, $\mathcal{I}_1,\mathcal{I}_2 \in \mathcal{I}_x$ specifies the input of the current node, $\mathcal{O}_1,\mathcal{O}_2 \in \mathcal{O}_x$ specifies the operation to be applied to the input tenor $\mathcal{I}_j$ of node $d$, and $\mathcal{M}$ specifies the element-wise addition operation that sums up the two operation's results of a node to generate the feature map corresponding to the output of this node, which is denoted by $\mathcal{P}_i^d$.

The set of possible inputs $\mathcal{I}_x$ is composed of the set of all previous nodes inside the block, (${\mathcal{P}_i^1,\mathcal{P}_i^2,......,\mathcal{P}_i^{d-1}}$), the output of the previous block, $\mathcal{P}_{i-1}$, the output of the previous-previous block, $\mathcal{P}_{i-2}$. Clearly, for each computation node, the input search space may change. $\mathcal{O}_x$ is the operation space consisting of the set of possible operations. Finally, all the outputs of the nodes that were not used as inputs to any other node within the block will be concatenated along the depth dimension to form the block's output.

\subsection{From block to neural network}

\begin{figure}[H]
\centering 
\includegraphics[height=3.0cm, width=0.48\textwidth]{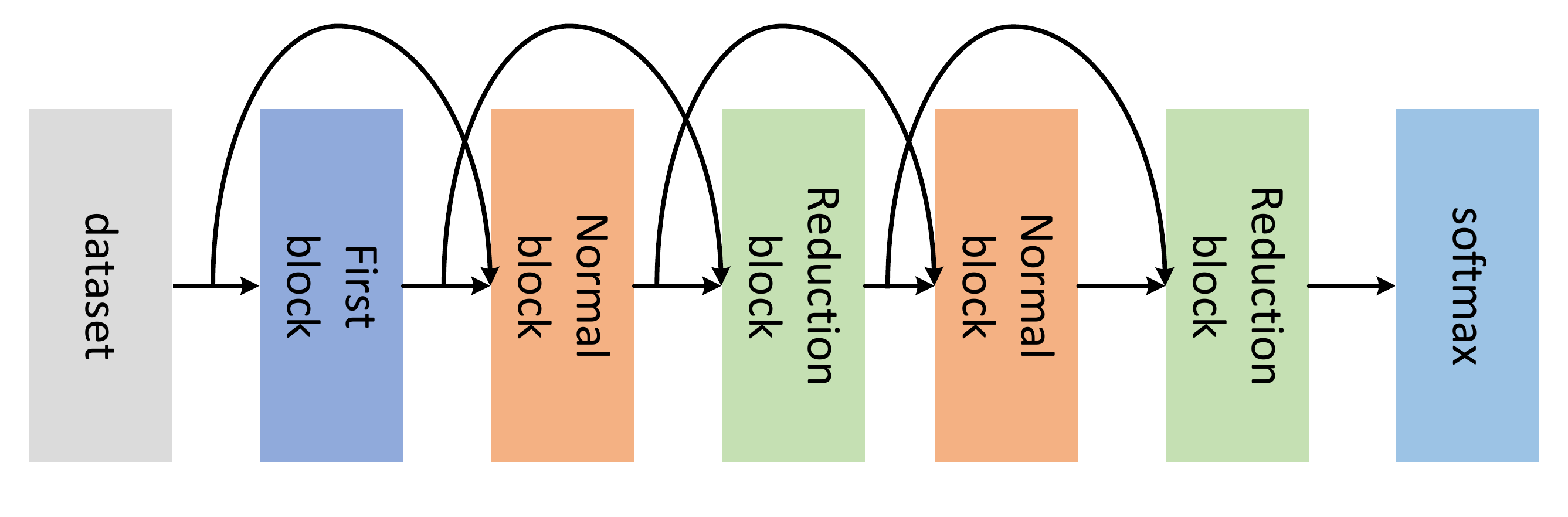} 
\caption{Connecting five blocks, the first block followed by two sets of the normal block and the reduction block, to form the overall network.} 
\label{Fig_3}
\end{figure}

In our work, three types of blocks can be designed by the proposed algorithm: the first block, the normal block, and the reduction block. Each block maps an $h \times w \times c$ tensor to another $H \times W \times C$ tensor. The only difference between the first block and other blocks is that the first block has one source node only. Moreover, the normal block applies all operations with a stride of 1, thus $H = h$ and $W = w$; the reduction block applies all operations with a stride of 2, thus $H = h/2$ and $W = w/2$. Hence, the reduction block can increase the receptive field of the deeper layers and reduce the spatial dimension of feature maps. As shown in Fig. \ref{Fig_3}, the network architecture begins with a first block, followed by two sets of blocks consisting of a normal block and a reduction block (with the same structure, but untied weights). All blocks are connected by a skip connection. At the end of the network, we utilize a softmax layer as the output layer of the neural network instead of large fully connected layers \cite{sun2018automatically}.

\subsection{Encoding strategy}

\begin{table}[]
\caption{The genotype-phenotype mappings}
\centering
\begin{tabular}{ccc}
\hline
\textbf{Phenotype}                                                                                     & Short name & Genotype \\ \hline
\textbf{identity mapping}                                                                              & Identity   & 1        \\ \hline
\textbf{\begin{tabular}[c]{@{}c@{}}depthwise separable\\  convolution with kernel size 3\end{tabular}} & DW3        & 2        \\ \hline
\textbf{\begin{tabular}[c]{@{}c@{}}depthwise separable \\ convolution with kernel size 5\end{tabular}} & DW5        & 3        \\ \hline
\textbf{FR Convolution with kernel size 3}                                                        & FR3        & 4        \\ \hline
\textbf{FR Convolution with kernel size 5}                                                        & FR5        & 5        \\ \hline
\textbf{Average pooling with kernel size 3}                                                            & AVG        & 6        \\ \hline
\textbf{max pooling with kernel size 3}                                                                & MAX        & 7        \\ \hline
\end{tabular}
\end{table}

The encoding strategy defines the genotype-phenotype, which is required for an EA to be employed to optimize the architecture of neural networks. Here, the phenotypes are different neural network architectures and the genotypes are the genetic encoding. The proposed encoding strategy aims at initializing a set of neural networks with different architectures by individuals in the EA. In our work, the EA only designs the computational nodes in each block.

\begin{figure}[H]
\centering 
\includegraphics[height=5.5cm, width=0.45\textwidth]{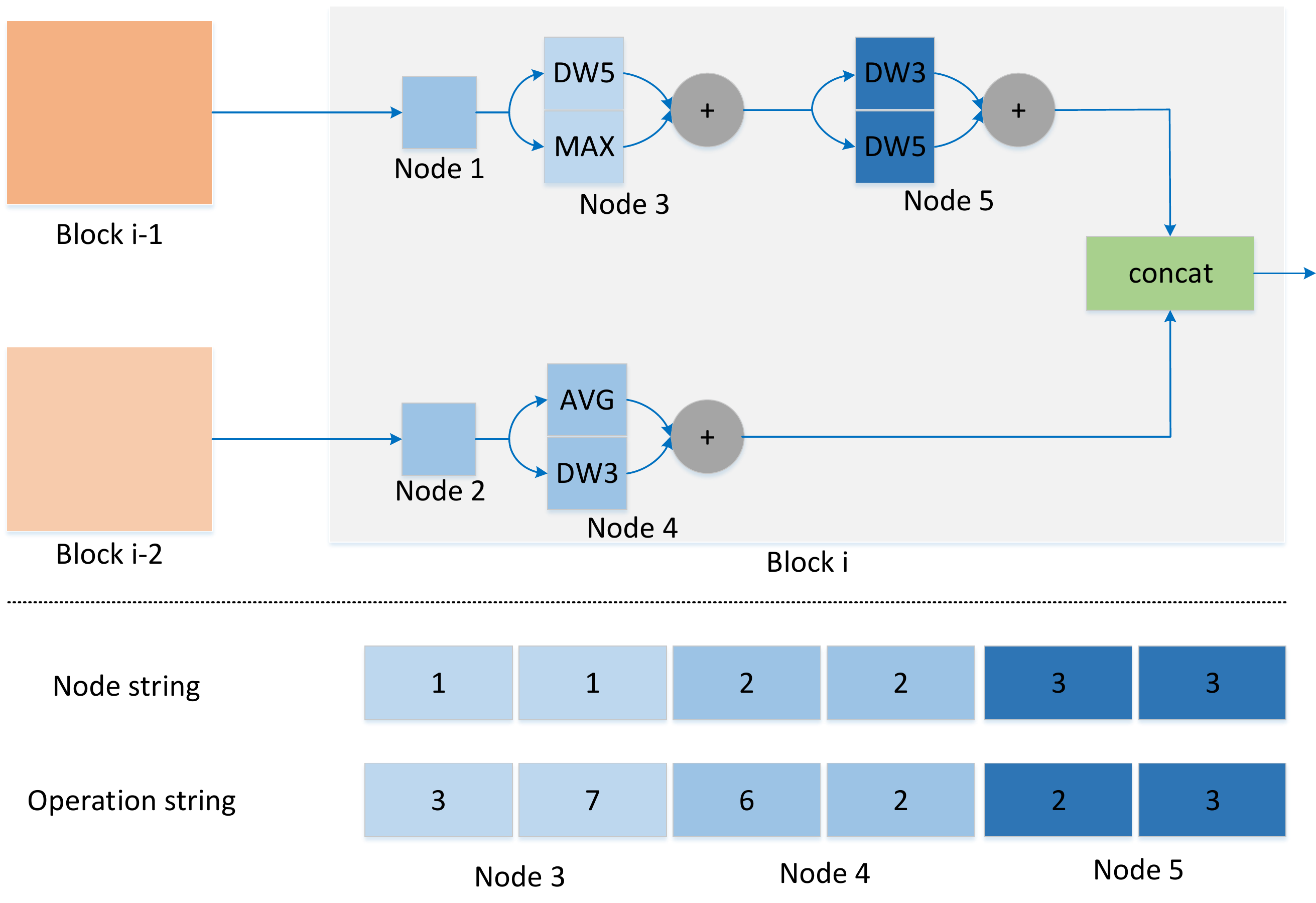} 
\caption{A block structure and its chromosome. The block has two source nodes and $N_c = 3$ computation nodes. In the figure, node\,1 and node\,2 receive their input from block i-1 and i-2. The first two integers 1,1 in the node string indicate that both operations of node 3 receive input from node 1. The first two integers 3,7 in the operation string denote that operations DW5 and MAX are applied to the two inputs, respectively.} 
\label{Fig_4}
\end{figure}

The chromosome for each block consists of a node string and an operation string. The node string represents the input of the corresponding node inside the block, while the operation string represents the operation type of each input. The chromosome is fully described by a tuple ($(\mathcal{I}_1,\mathcal{I}_2,......),(\mathcal{O}_1,\mathcal{O}_2,......)$). All genotype-phenotype mappings used in this work are presented in Table I. In the table, DW is a depth-wise separable and efficient convolution operation. Depth-wise separable convolution is able to reduce network parameters without losing network performance. Here we use two DW operations with a kernel size $3\times3$ and $5\times5$, SW3 and SW5 for short. In addition, FR represents a feature reconstruction convolutional operation, which consists of a normal convolutional layer followed by a channel attention module. In this work, two available operators are FR3 and FR5, which have a normal convolutional layer with kernel size $3\times3$ and $5\times5$, respectively. More discussions about the FR operation will be discussed in Section III.D. Fig. \ref{Fig_4} provides an example of an encoded block and the corresponding network structure, where $N_c=3$.

\subsection{Multi-scale feature reconstruction convolutional operation}

Inspired by the previous work on efficient channel attention mechanism \cite{wang2019eca-net:}, we propose a feature reconstruction convolutional operation as a building block in the neural architecture search, FR for short. FR consists of a normal convolutional layer followed by a channel attention module.

\begin{figure}[H]
\centering 
\includegraphics[height=3.2cm, width=0.45\textwidth]{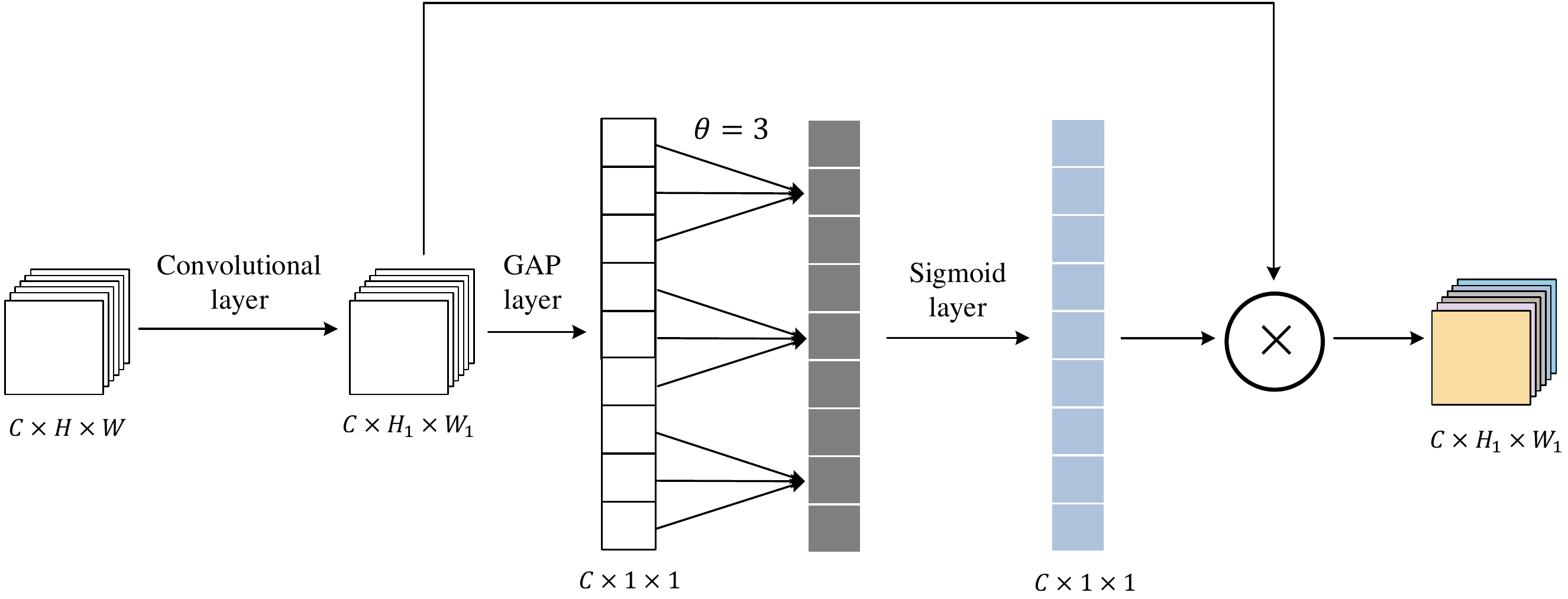} 
\caption{The structure of the FR convolutional operation. An array of the square block represents the input features. After a global average pooling (GAP) layer is the channel attention module followed by a $\sigma_{sigmoid}$ function layer that computes weights using a $1D$ convolution of size $\theta$. The rectangles represent feature maps of channels. Gray rectangles represent the output of $1D$ convolution. Blue rectangles represent the weight of channels. An array of the colorful square block represents the output feature maps of FR operation.} 
\label{Fig_5}
\end{figure}

Fig. \ref{Fig_5} plots the components of the FR convolutional operation. Specifically, given the input features, FR first uses a normal convolutional layer to extract the features. Different convolutional kernel sizes can be used to capture the spatial information at different scales (both at fine and coarse grain level) \cite{khan2019survey}. The normal convolutional layer expands the input feature maps from $x\in {\mathbb{R}}^{C\times H\times W}$ to feature maps $T(x)\in {\mathbb{R}}^{C\times H_1\times W_1}$. Then a global average pooling (GAP) layer is employed for each channel independently. After that, a $1D$ convolutional layer of size $\theta$ followed by a $\sigma_{sigmoid}$ function layer is utilized to generate the weight of each channel. The $1D$ convolutional layer of size $\theta$ is designed to capture the non-linear cross-channel interaction for each channel with its $\theta$ neighboring channels, where the kernel size $\theta$ represents how many neighboring channels take part in the attention prediction of the channel. Parameter $\theta$ is adaptively determined by an exponential function proposed in \cite{wang2019eca-net:}.

Channel attention is generated by the feedforward process and learned by the feedback process. The whole structure can be trained end-to-end. During the feedforward process, FR is able to reconstruct the channel-wise feature response to reduce the feature redundancy of channels. During the back-propagation process, FR can prevent unimportant gradient (from the unimportant channel) to update the parameter \cite{wang2017residual}.

\section{Proposed algorithm}

As discussed above, the search space of the proposed SI-ENAS is represented using a single DAG, where a neural network architecture can be realized by taking a subgraph of the DAG. Different connection relationships between the nodes will result in a large number of neural networks with different architectures. We use SI-ENAS to learn the connection relationship between nodes and to find better topologies for deep neural networks. Except for the source nodes, each node in DAG represents some local computation, which is specified by the weights and bias. Since an offspring individual (a new neural network model) generated by applying genetic operations on parent individuals (existing neural network models in the parent population) can be seen as the recombination of the nodes of the parent models, the parameters of the offspring individual can be directly inherited from the parent networks. We call this method node inheritance. Moreover, because each node is repeatedly used throughout the evolutionary optimization, all parameters of the nodes are trained and updated in the evolutionary search process. Consequently, SI-ENAS is able to avoid training offspring individuals from scratch with the help of node inheritance, thereby effectively reducing the high computation costs usually required by ENAS.

\subsection{Overall framework}

\begin{figure*}
\centering 
\includegraphics[height=10cm, width=1\textwidth]{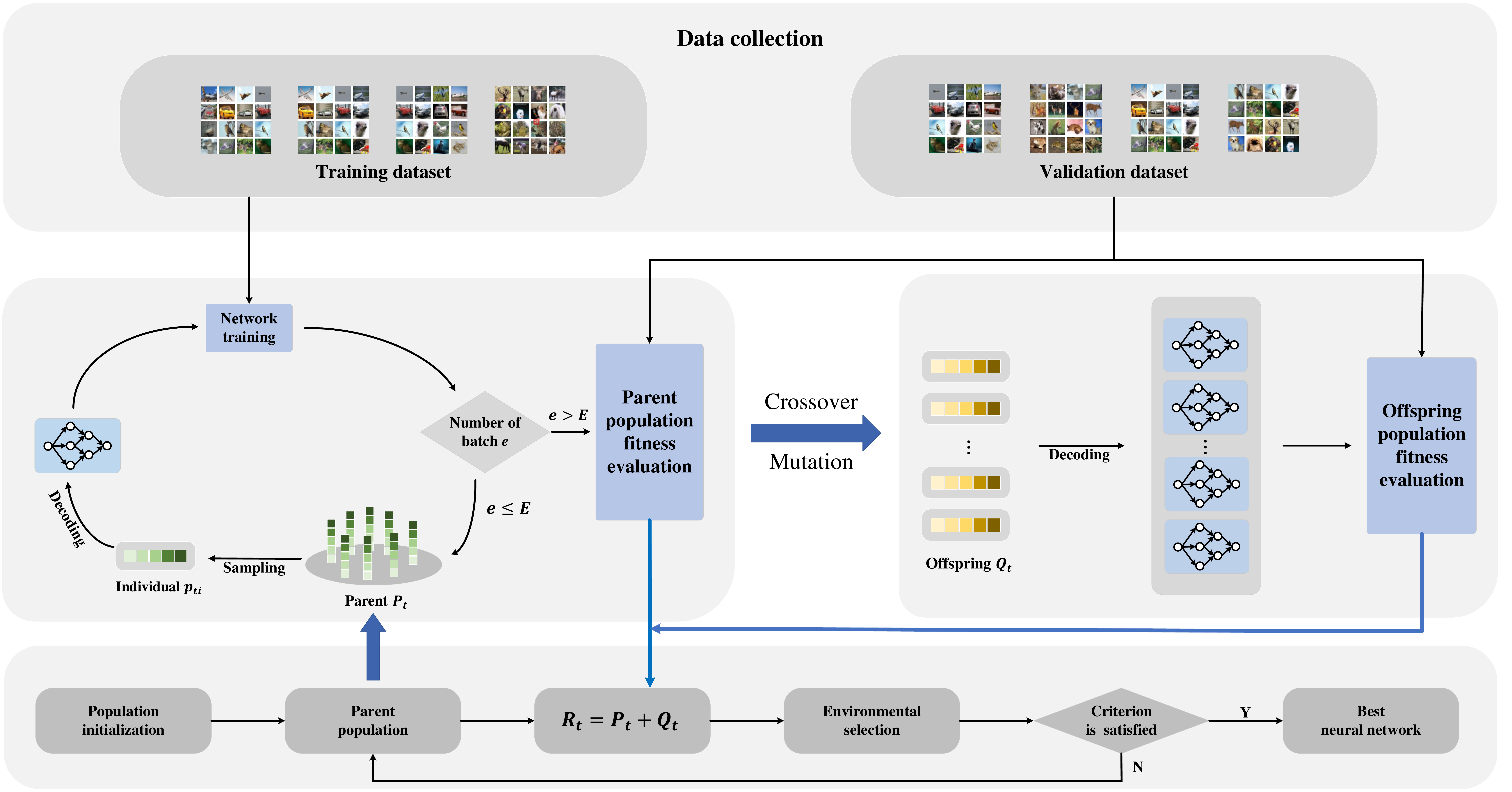} 
\caption{The overall framework of SI-ENAS} 
\label{Fig_6}
\end{figure*}

\begin{algorithm}[htbp]\footnotesize{
\caption{The framework of SI-ENAS} \algblock{Begin}{End}
\label{Algorithm 1}
\renewcommand{\algorithmicrequire}{\textbf{Input:}}
\renewcommand{\algorithmicensure}{\textbf{Output:}}
\begin{algorithmic}[1]
 \Require
 population size $K$, maximum number of generations $G$, batch size $b$, training data $D_{train}$, validation dataset $D_{valid}$
 \Ensure
 the best neural network architecture
\State $P_0$ $\leftarrow$ initialize a population with a size of $K$ by Algorithm 2
\State $t$ $\leftarrow$ 0
\While{$t < G$}
\For {each $batch$ data in $D_{train}$}
    \State $net$ $\leftarrow$ randomly sample an individual from $P_t$ and decode it into the corresponding neural network
    \State Training $net$ on mini-batch $D_{train}$
\EndFor
\For {each $individual$  in $P_t$}
    \State Calculate the classification accuracy of $individual$ on $D_{valid}$
\EndFor
    \State generate offspring $|{Q_t}|=K$ by Algorithm 3
\For {each $individual$  in $Q_t$}
    \State Calculate the classification accuracy of $individual$ on $D_{valid}$
\EndFor
    \State $R_t$ $\leftarrow$ $P_t$ $\cup$ $Q_t$
    \State $p_{best}$,${\delta}_{best}$ $\leftarrow$ select the individual with the best fitness from $R_t$
    \State Set ${\delta}_{best}$ as the fitness of $R_t$
    \State $P_{t+1}$ $\leftarrow$ select $K$ individuals from $R_t$ by environment selection
    \If {$p_{best}$ is not in $P_{t+1}$}
        \State Replace the one that is the worst individual in $P_{t+1}$ by $p_{best}$
    \EndIf
    \State $t$ $\leftarrow$ $t+1$
\EndWhile
\State Return the best individual from $P_t$ and decode it into the corresponding neural network
\end{algorithmic}}
\end{algorithm}

$\mathbf{Algorithm\,1}$ lists the main components of the SI-ENAS. It starts with an initial population of $P_0$ consisting of $K$ randomly generated individuals (line 1). Then, we repeat the following steps (lines 4-22) for $G$ generations. Each generation is composed of training and evaluation of parent individuals (line 4-10), generation and evaluation of offspring individuals (line 11-14), combination of the parent and offspring populations (line 15), and environmental selection with an elitism strategy (line 16-21). The main differences between SI-ENAS and most existing ENAS lie in population initialization, parent population training and fitness evaluation of the offspring using node inheritance, which are to be detailed in Subsections IV-B and IV-C.

In SI-ENAS, the fitness of an individual is the classification accuracy of the neural network decoded from the individual on the given validation dataset. When decoding a neural network, we need to pay attention to the following practices of modern CNNs \cite{he2016deep}. First, a batch normalization operation followed by a ReLU activation function is added to the output of the depth-wise separable convolution layer. Second, the pooling layer starts with the ReLU activation function. Third, the zero-padding operation is used to make the size of the input feature map and the output feature map of each node to be the same.

The selection operation plays a key role in enhancing the performance of SI-ENAS, including mate selection (selection of parent individuals for reproduction) and environmental selection (selection individuals for the parents of the next generation).  In mate selection, the binary tournament selection is adopted that randomly picks two individuals from the parent population and selects the one with the better fitness. In environmental selection, a population of individuals with a size of $K$ is selected from the combined population, $R_t$, as the parent individuals for the next generation. Theoretically, in order to prevent the search from getting trapped in a local minimum and to avoid premature convergence, a sufficient degree of population diversity should be maintained \cite{back1996evolutionary}. In our algorithm, we select the individual by the binary tournament selection to enhance the diversity of the population. However, the best individual in the parent population may be lost using the tournament selection. Hence, we always pass the best individual into the next population, which is called an elitism strategy in EAs.

The framework of the overall SI-ENAS is illustrated in Fig. \ref{Fig_6}. Note that in SI-ENAS, each individual in the parent population is trained by a sampling method. That is, for each mini-batch of the training data, one individual in the parent population is randomly selected and trained on $D_{train}$. When the training of the parent population is finished, all parameter of computation nodes in search space are updated. After that, all parent individuals are tested on the validation dataset to calculate their fitness.

Subsequently, an offspring population are generated by means of node inheritance, which will be detailed in Subsection IV.C. Then, the offspring individuals will be directly tested on the validation dataset without training. When the evolutionary loop terminates, SI-ENAS outputs the best individual and decodes it into the corresponding neural network architecture. It should be pointed out that the best individual will undergo a complete training before it is tested on the test dataset for a final performance assessment.

From the above discussions, we can see that the proposed fast ENAS framework, SI-ENAS, is significantly different from the conventional ENAS shown in Fig. \ref{Fig_1}. The main differences are that in SI-ENAS, the parent individuals are randomly sampled and trained on mini-batches of training data, while the offspring individuals generated using a node inheritance strategy and do not need to be trained for fitness evaluations. This way, SI-ENAS is able to significantly reduce the training time while avoiding biases towards either parent or offspring individuals.

\subsection{Population initiation}

\begin{algorithm}[htbp]\footnotesize{
\caption{Population initialization} \algblock{Begin}{End}
\label{Algorithm 2}
\renewcommand{\algorithmicrequire}{\textbf{Input:}}
\renewcommand{\algorithmicensure}{\textbf{Output:}}
\begin{algorithmic}[1]
 \Require
 population size $K$, $node\,space$, $operation \,space$
 \Ensure
 the initial population $P_0$
\State $P_0$ $\leftarrow$ $\varnothing$
\State $first\,block$,$normal\,block$,$reduction\,block$ $\leftarrow$ Create three types of blocks containing $N_C$ nodes
\While{$ |P_0 |< K$}
\State $node\,string$  $\leftarrow$ $\varnothing$
\State $operation\,string$  $\leftarrow$ $\varnothing$
\For {each computation node in $first\, block$}
    \State $\mathcal{I}_{f1}$,$\mathcal{I}_{f2}$ $\leftarrow$  randomly select two integers from $node\,space$
    \State $\mathcal{O}_{f1}$,$\mathcal{O}_{f2}$ $\leftarrow$  randomly select two integers from $operation\,space$
    \State $node\,string$ $\leftarrow$ $node\,string$ $\cup$ $\mathcal{I}_{f1}$,$\mathcal{I}_{f2}$
    \State $operation\,string$ $\leftarrow$ $operation\,string$ $\cup$  $\mathcal{O}_{f1}$,$\mathcal{O}_{f2}$
\EndFor
\State $first\,block$ $\leftarrow$ $node\,string$ $\cup$ $operation\,string$
\For {each computation node in $normal\, block$}
    \State $\mathcal{I}_{n1}$,$\mathcal{I}_{n2}$ $\leftarrow$ randomly select two integers from $node\,space$
    \State $\mathcal{O}_{n1}$,$\mathcal{O}_{n2}$ $\leftarrow$ randomly select two integers from $operation\,space$
    \State $node\,string$ $\leftarrow$ $node\,string$ $\cup$ $\mathcal{I}_{n1}$,$\mathcal{I}_{n2}$
    \State $operation\,string$ $\leftarrow$ $operation\,string$ $\cup$ $\mathcal{O}_{n1}$,$\mathcal{O}_{n2}$
\EndFor
\State $normal\, block$ $\leftarrow$ $node\,string$ $\cup$ $operation\,string$
\For {each computation node in $reduction\, block$}
    \State $\mathcal{I}_{r1}$,$\mathcal{I}_{r2}$ $\leftarrow$  randomly select two integers from $node\,space$
    \State $\mathcal{O}_{r1}$,$\mathcal{O}_{r2}$ $\leftarrow$  randomly select two integers from $operation\,space$
    \State $node\,string$ $\leftarrow$ $node\,string$ $\cup$ $\mathcal{I}_{r1}$,$\mathcal{I}_{r2}$
    \State $operation\,string$ $\leftarrow$ $operation\,string$ $\cup$ $\mathcal{O}_{r1}$,$\mathcal{O}_{r2}$
\EndFor
\State $reduction\, block$ $\leftarrow$ $node\,string$ $\cup$ $operation\,string$
\State $P_0$ $\leftarrow$  $first\,block$ $\cup$ $normal\,block$ $\cup$ $reduction\,block$
\EndWhile
\State return $P_0$
\end{algorithmic}}
\end{algorithm}

As introduced in Section III.B, we encode a neural network by a chromosome of 2-tuple [$node\, string$, $operation\, string$], where $node\, string$ represents the input of each corresponding node and $operation\, string$ represents the operation applied to the input. The details of population initialization are described in $\mathbf{Algorithm\,2}$. For each computation node in the block, the EA first selects the two integers from the input search space $node\, space$ as the input to the current node. Then, two integers are selected from the operation space $operation\,space$ as the corresponding operations. This process repeats until all nodes are configured. After that, the nodes are linked and stored in $P_0$. Finally, $K$ individuals are randomly initialized in the same way and they are stored in $P_0$.

\subsection{Node inheritance}

\begin{figure*}
\centering 
\includegraphics[width=0.8\textwidth]{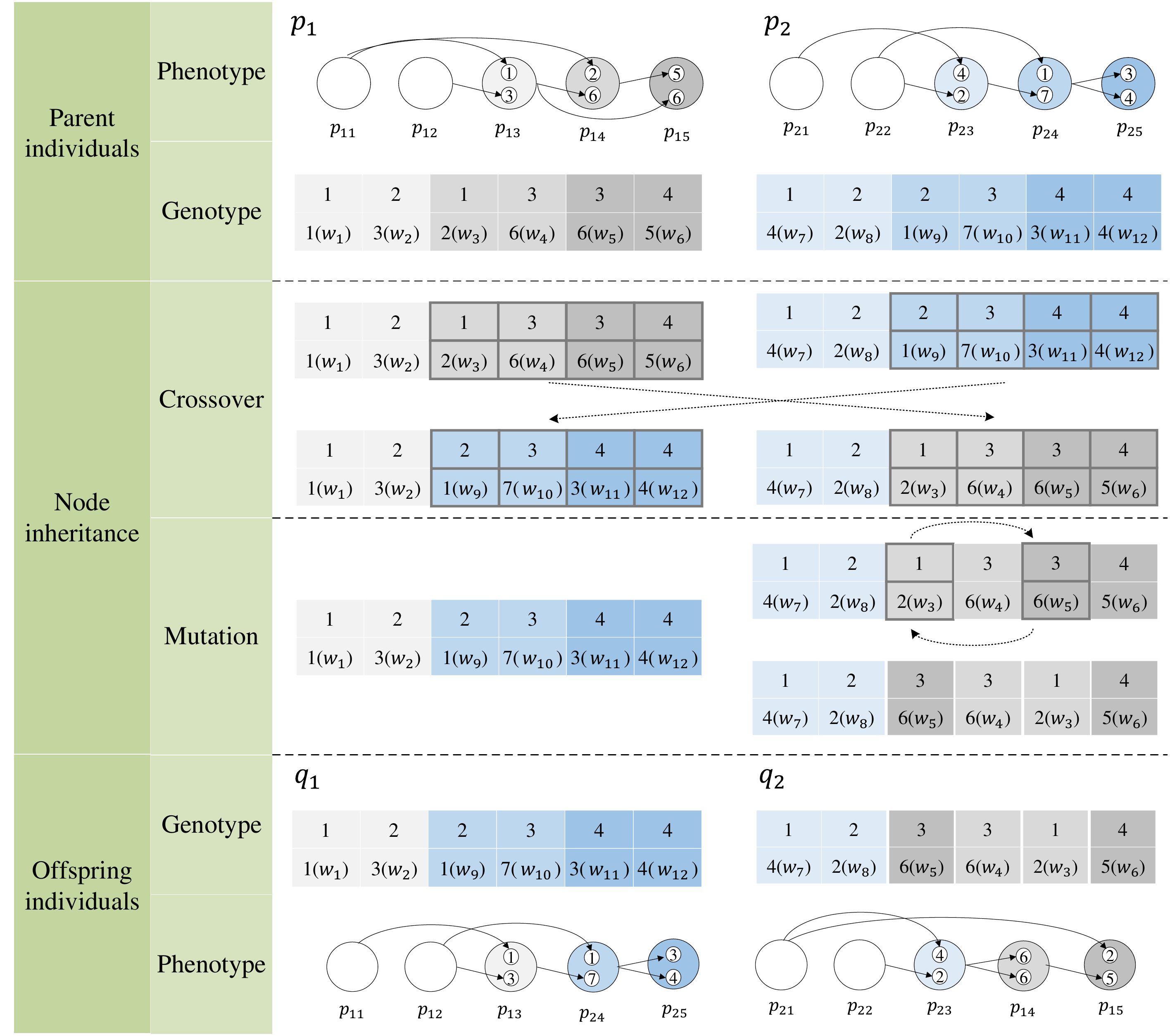} 
\caption{An example of offspring generation by means of node inheritance, including crossover, exchange mutation and weight inheritance. Given two parents, $p_1$ and $p_2$, two offspring individuals, $q_1$ and $q_2$ are created by a crossover between the second gene and third gene. Then, one of of the offspring $q_2$ has a mutation by exchanging the third gene and the fifth gene. All weights are directly inherited from their parents.} 
\label{Fig_7}
\end{figure*}

By node inheritance, we mean in this work to generate new neural network architectures (offspring) from parents using crossover, exchange mutation and weight inheritance. Both node/operation crossover and mutation can be achieved by exchanging the access order of the computing nodes in the DAGs. Note that no untrained new node will be generated in node inheritance and consequently, no weight training of the offspring individuals is needed before evaluating their fitness on the validation dataset.

$\mathbf{Algorithm\,3}$ shows the details of the node inheritance operator in SI-ENAS. The first part is one-point crossover(lines 3-12). Two parents are selected from the population $P_t$ using the binary tournament selection to create two offspring by applying one-point crossover on the node and operation strings. This process is repeated until $K$ offspring individuals are generated. The second part is the node/operation exchange mutation (lines 14-19), which can be seen as a type of mutation in which an individual randomly exchanges the order of two computation nodes or operations in its chromosome.

Fig. \ref{Fig_7} provides an illustrative example of offspring generation by means of node inheritance. In the example, two parents, $p_1$ and $p_2$ are chosen using the tournament selection for reproduction. The position between the second and third genes are chosen as the crossover point to apply one-point crossover. Then, an exchange mutation occurs to $q_2$, in which its third gene and fifth gene exchange their position. This way, all nodes, including their weights, in the offspring are inherited from the parents and therefore, no training is required for fitness evaluation. Although one-point crossover and  exchange mutation are simple, they have been shown to considerably improve the performance of NAS, which will be experimentally demonstrated in Section VI.

\begin{algorithm}[htbp]\footnotesize{
\caption{Node inheritance} \algblock{Begin}{End}
\label{Algorithm 3}
\renewcommand{\algorithmicrequire}{\textbf{Input:}}
\renewcommand{\algorithmicensure}{\textbf{Output:}}
\begin{algorithmic}[1]
 \Require
  population $P_t$, probability of crossover $P_c$, probability of mutation $P_m$.
 \Ensure
 offspring population $Q_t$
\State $Q_t$ $\leftarrow$ $\varnothing$
\While {$|Q_t| < |P_t|$}
    \State $p_1$,$p_2$ $\leftarrow$ select two individuals from population $P_t$ by the binary tournament selection
    \State $\gamma$ $\leftarrow$ uniformly generate a number from (0,1]
    \If {$\gamma<P_c$}
    \State $q_1$,$q_2$ $\leftarrow$ generate offspring by one-point crossover
    \Else
    \State $q_1$ $\leftarrow$ $p_1$
    \State $q_2$ $\leftarrow$ $p_2$
    \EndIf
    \State $Q_t$ $\leftarrow$ $Q_t$ $\cup$ $q_1$ $\cup$ $q_2$
\EndWhile
\For {each individual $p$ in $Q_t$}
    \State $\gamma$ $\leftarrow$ uniformly generate a number from (0,1]
    \If {$\gamma<p_m$}
    \State $q_n$ $\leftarrow$ change the individual by exchange mutation
    \EndIf
\EndFor
\State return $Q_t$
\end{algorithmic}}
\end{algorithm}

\section{Experimental settings}

The final goal of SI-ENAS is to efficiently find the optimal neural network architecture which achieves promising classification accuracy based on the benchmark dataset. To this end, a series of experiments are designed in this work to demonstrate the advantage of the proposed approach compared to the state-of-the-art. First, we evaluate the performance of the proposed algorithm by investigating the classification performance of the evolved neural networks. Second, we examine the effectiveness of the proposed node inheritance and FR operation. Finally, we transfer the optimized network architecture evolved on CIFAR10 to CIFAR100 and SVHN to evaluate the transferability of the evolved network architecture.

In this section, the peer competitors chosen to compare with the proposed algorithm are introduced in Subsection V.A. Then, the used benchmark datasets are to be introduce in Subsection V.B. Finally, the parameter settings of the SI-ENAS and the final test if the evolved neural network are presented in Subsection V.C.

\subsection{Peer competitors}

In order to demonstrate the superiority of the proposed algorithm, various peer competitors are selected for comparison. The selected competitors can be divided into three different groups.

The first group includes the state-of-the-art CNN architectures which are manually designed by human experts, including DenseNet\cite{huang2017densely}, ResNet \cite{he2016deep}, Pre-act-ResNet-110 \cite{he2016identity}, Maxout \cite{goodfellow2013maxout}, VGG \cite{simonyan2014very}, Network in network \cite{lin2013network}, Highway network \cite{srivastava2015highway}, All-CNN \cite{springenberg2014striving}, Fractal-Net \cite{larsson2016fractalnet}, DSN \cite{zagoruyko2016wide}, Residual-attention-236 \cite{wang2017residual} IGCV3-D ($G_1=4$, $G_2=2$) \cite{sun2018igcv3}. Considering the attractive performance of ResNet, we utilize three different network depths of 56, 101, 1202.

The second group comprises various non-evolutionary NAS methods, such as NAS \cite{zoph2016neural}, MetaQNN \cite{baker2016designing}, EAS \cite{zoph2016neural}, and Block-QNN-S \cite{zhong2017practical}.

The third group represents the state-of-the-art evolutionary NAS for CNN architecture design, including Genetic-CNN \cite{xie2017genetic}, neural networks evolution \cite{zhang2018finding}, Large-scale Evolution \cite{real2017large-scale}, Hierarchical evolution \cite{liu2017hierarchical}, AmoebaNet + cutout \cite{real2019regularized}, CGP-CNN \cite{suganuma2017genetic}, CoDeepNEAT \cite{miikkulainen2019evolving}, CNN-GA \cite{sun2018automatically}, AE-CNN \cite{sun2018automatically}, and AE-CNN+E2EPP \cite{sun2019surrogate}.

\subsection{Benchmark datasets}

In our experiments, the benchmark datasets include CIFAR10 \cite{krizhevsky2009learning}, CIFAR100 \cite{krizhevsky2009learning}, and SVHN datasets \cite{netzer2011reading}, which are all widely adopted in state-of-the-art CNNs and NAS algorithms.

CIFAR10 is a 10-category classification dataset consisting of 50,000 training datasets and 10,000 test datasets, and each image has a dimension of $32 \times 32$. CIFAR100 has the same number of images in the training datasets and test datasets as those of CIFAR10, except that it is 100-category classification problem.

The SVHN (Street View House Number) dataset is composed of 630,420, $32 \times 32$ RGB color images in total, of which 73,257 samples are used for training, 26,032 for validation, and the rest 531,131 for test.  The task of this dataset is to classify the digit located at the center of each image.

In the experiments, to avoid seeing the test data in the evolutionary process, the training datasets of each benchmark dataset are divided into two parts. The first part accounts for $80\%$ of the data and is used as the training dataset. The remaining  $20\%$  images are used for validation in calculating the fitness value. Moreover, the datasets are processed by the same data pre-processing and augmentation routine, which is often used in peer competitors during the training \cite{huang2017densely, he2016deep}.

\subsection{Parameter settings}

In this subsection, the parameter settings for SI-ENAS are detailed. All the parameter settings are summarized in Table II. The parameter settings are applied to all experiments.

Our algorithm parameters are divided into two parts: evolutionary search and best individual validation. For the evolutionary search, the parameter settings follow the practices in evolutionary computation. Although a larger population size and a larger number of generations will in principal lead to better performance, the computation costs will also become prohibitive. Hence, we investigate the impact of the population size and the maximum number of generations on the performance and computation cost of SI-ENAS in Subsection V-D. The probabilities of crossover and mutation are set to 0.95 and 0.05, respectively. The mini-batch SGD is used to train both individuals whose weights are initialized with the He initialization \cite{he2015delving}. Meanwhile, the momentum, nesterov, weight decay, drop out are adopted as widely accepted in the deep learning community. During the evolutionary search, we run the evolutionary process for 300 generations and set the population size to 25. The learning rate is set to 0.1 from the first generation to the 149-th generation, 0.01 from the 150-th generation to the 224-th generation, and 0.001 for the remaining generation.

When the evolutionary process terminates, the best individual is retrained from scratch and its classification accuracy is evaluated on the validation dataset. In this training, the best neural network is trained 500 epochs, and the learning rate is initialized to 0.05 and scaled by dividing it by 10 at the 300-th epoch and again at the 450-th epoch. The other parameters are set the same as in the evolutionary search. Finally, the trained neural network is tested on the test dataset. The test classification accuracy is reported as the best result of our experiments, following the practice in deep learning. Note that these experimental settings are constrained by the computational resources available to us. All experiments are performed on one Nvidia GeForce RTX 2080Ti.

\begin{table}[]
\caption{A summary of the parameter settings}
\begin{tabular}{c|c|c}
\hline
\multirow{12}*{SI-ENAS}& Parameter name            & Parameter value\\  \hline
 & Computation node $N_c$     & 5                                                                     \\
 \cline{2-3}
 & Crossover probability     & 0.95                                                                     \\
 \cline{2-3}
 & Mutation probability      & 0.05                                                                     \\
 \cline{2-3}
 & Batch size                & \begin{tabular}[c]{@{}c@{}}128(CIFAR10, SVHN)\\64(CIFAR100)\end{tabular}                                     \\
 \cline{2-3}
 & Weight decay              & $1 \times 10^{-4} $                                                        \\
 \cline{2-3}
 & Initialized Learning rate & 0.1                                                                      \\
 \cline{2-3}
 & Optimizer                 & SGD                                                                      \\
 \cline{2-3}
 & Momentum                  & 0.9                                                                      \\
 \cline{2-3}
 & Nesterov                  & Ture                                                                     \\
 \cline{2-3}
 & Dropout                   & 0.5                                                                      \\
 \cline{2-3}
 & Channel number            & \begin{tabular}[c]{@{}c@{}}32(CIFAR10, SVHN)\\ 64(CIFAR100)\end{tabular} \\ \hline

 Best individual     & Initialized Learning rate & 0.05                       \\
 \cline{2-3}
 validation     & Epoch                     & 500                                                                      \\ \hline
\end{tabular}
\end{table}

\subsection{Sensitivity to population size}

\begin{figure}
\begin{minipage}[t]{0.5\linewidth}
\centering
\includegraphics[width=1.55in]{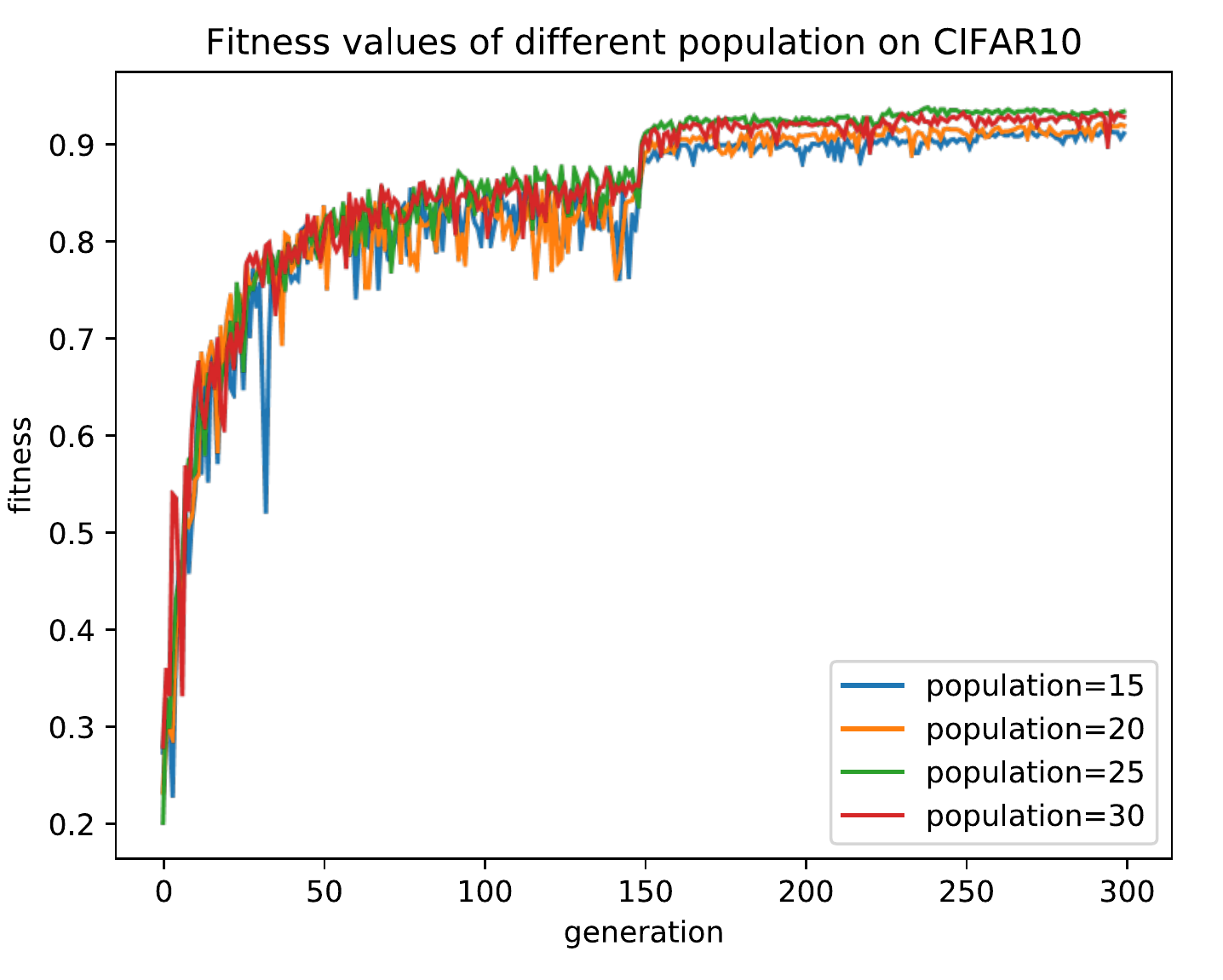}
\end{minipage}%
\begin{minipage}[t]{0.5\linewidth}
\centering
\includegraphics[width=1.60in]{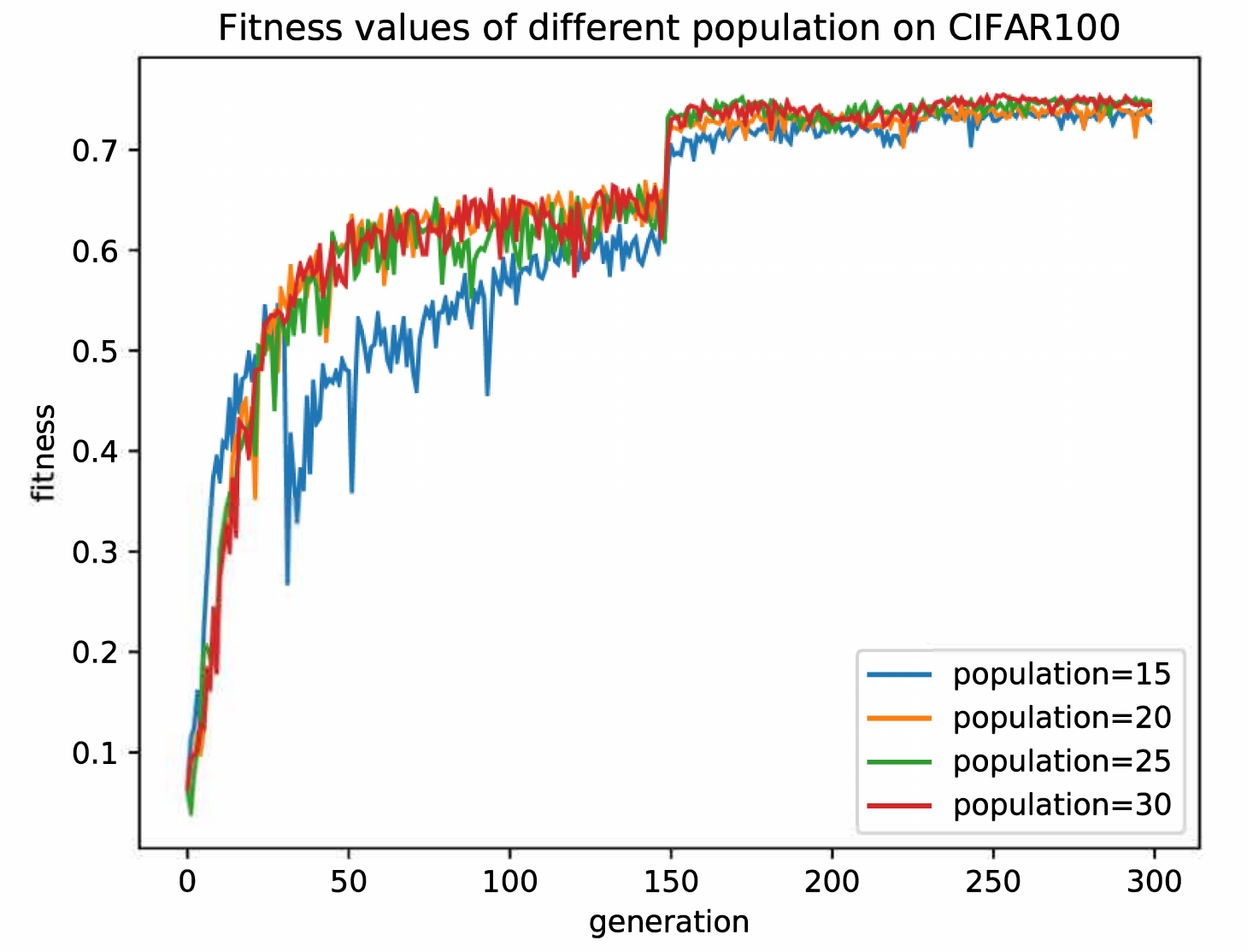}
\end{minipage}
\caption{Change of best fitness value of the population over the generations for different population sizes (15, 20, 25, 30) on the CIFAR10 (left) and CIFAR100 (right) datasets.} 
\label{Fig_8}  
\end{figure}

To investigate the influence of the population size on the performance and computation cost of the proposed algorithm, we set the population size $K=15, 20, 25, 30$ and run the evolutionary search for 300 generations on CIFAR10 and CIFAR100. The change of the classification accuracy of the best individual in the population over the generations for different population sizes are plotted in Fig. \ref{Fig_8}. The final classification accuracy and runtime of SI-ENAS with different population sizes are listed in Table III, where the runtime is in GPU days, which is a unit proposed in \cite{sun2019surrogate} meaning a single GPU is fully utilized in one day.  From these results, we can conclude that the best performance (93.7\% on CIFAR10 and 75.2\% CIFAR100) is achieved when the population size is set both to 25.  When the population size is increased to 30, no  performance improvement is observed, although the runtime will become higher and higher. Hence, the population size is set to 25 in the remaining experiments.

\begin{table}[]
\caption{The accuracy of the best network found by SI-ENAS and its runtime in GPU days on the CIFAR10 and CIFAR100 benchmark datasets when the population size is set to 15, 20, 25, and 30. }
\centering
\begin{tabular}{cccc}
\hline
\begin{tabular}[c]{@{}c@{}}Search dataset\end{tabular} & \begin{tabular}[c]{@{}c@{}}Population size\end{tabular} & \begin{tabular}[c]{@{}c@{}}Fitness value \\ of population\end{tabular} & \begin{tabular}[c]{@{}c@{}}Search time\\ (GPU days)\end{tabular} \\ \hline
                                                         & 15                                                        & 90.4                                                                   & 0.375                                                            \\
                                                         & 20                                                        & 92.4                                                                   & 0.459                                                            \\
CIFAR10                                                  & 25                                                        & 93.7                                                                   & 0.551                                                            \\
                                                         & 30                                                        & 93.5                                                                   & 0.646                                                            \\ \hline
                                                         & 15                                                        & 73.6                                                                   & 0.729                                                            \\
                                                         & 20                                                        & 74.4                                                                   & 0.792                                                            \\
CIFAR100                                                 & 25                                                        & 75.2                                                                   & 0.917                                                            \\
                                                         & 30                                                        & 75.1                                                                   & 1.042                                                            \\ \hline
\end{tabular}
\end{table}

\section{Comparative studies}

Here we conduct a series of comparative studies to demonstrate the advantage of the proposed algorithm. Subsection VI.A compared the proposed algorithm, SI-ENAS with 29 state-of-the-art NAS methods in terms of both classification accuracy and computation costs. In Subsections VI.B and VI.C, the effectiveness of FR operation and node inheritance are examined, respectively. Finally, we test the accuracy of the best architecture evolved on CIFAR10 on two different datasets, CIFAR100 and SVHN to investigate the transferability of the evolved neural architecture in Subsection VI.D.

\subsection{Overall results}
\begin{table*}[]
 \centering
 \caption{Comparison of SI-ENAS and the peer competitors in terms of the classification accuracy (\%) and the consumed GPU days on the CIFAR10 and CIFAR100 datasets}
\begin{tabular}{c|c|c|c|c|c|c}
\hline
\multirow{2}{*}{Method}                                                 & \multirow{2}{*}{Peer Competitors}                                                                                                                                                                                                                                                & \multirow{2}{*}{CIFAR10}                                                                                                                             & \multirow{2}{*}{CIFAR100}                                                                                                                       & \multirow{2}{*}{GPU Days}                                                                                   & \multicolumn{2}{c}{Performance Enhancement}                                                                                                                                                                                                                                      \\ \cline{6-7}
                                                                        &                                                                                                                                                                                                                                                                                  &                                                                                                                                                      &                                                                                                                                                 &                                                                                                             & CIFAR10                                                                                                                               & CIFAR100                                                                                                                                  \\ \hline
\begin{tabular}[c]{@{}c@{}}State-of-art CNN\\ architecture\end{tabular} & \begin{tabular}[c]{@{}c@{}}DenseNet(L = 40; k = 12) \cite{huang2017densely} \\ ResNet(depth=56) \cite{he2016deep} \\ ResNet(depth=101) \cite{he2016deep} \\ ResNet(depth=1202)\cite{he2016deep} \\ Pre-act-ResNet-110 \cite{he2016identity}\\ Maxout \cite{goodfellow2013maxout} \\ VGG \cite{simonyan2014very} \\ Network in Network \cite{lin2013network}\\ Highways network \cite{srivastava2015highway} \\ All-CNN \cite{springenberg2014striving} \\ Fractalnet \cite{larsson2016fractalnet} \\ DSN \cite{zagoruyko2016wide} \\ Residual-attention-236 \cite{wang2017residual} \\ IGCV3-D \cite{sun2018igcv3} \end{tabular} & \begin{tabular}[c]{@{}c@{}}94.76\\ 93.03\\ 93.57\\ 92.07\\ 93.63\\ 90.70\\ 93.34\\ 91.19\\ 92.28\\ 92.75\\ 94.78\\ 91.8\\ 95.86\\ 94.71\end{tabular} & \begin{tabular}[c]{@{}c@{}}75.58\\ --\\ 74.84\\ 72.18\\ --\\ 61.40\\ 67.95\\ 64.32\\ 67.61\\ 66.29\\ 77.70\\ 64.30\\ 78.84\\ 77.84\end{tabular} & \begin{tabular}[c]{@{}c@{}}--\\ --\\ --\\ --\\ --\\ --\\ --\\ --\\ --\\ --\\ --\\ --\\ --\\ --\end{tabular} & \begin{tabular}[c]{@{}c@{}}1.17\\ 2.9\\ 2.36\\ 3.86\\ 2.3\\ 5.23\\ 2.59\\ 4.74\\ 3.65\\ 3.18\\ 1.15\\ 4.13\\ 0.07\\ 1.22\end{tabular} & \begin{tabular}[c]{@{}c@{}}5.78\\ --\\ 6.52\\ 9.18\\ --\\ 19.96\\ 13.41\\ 17.04\\ 13.75\\ 15.07\\ 3.66\\ 17.06\\ 2.52\\ 3.52\end{tabular} \\ \hline
\begin{tabular}[c]{@{}c@{}}Non-evolutionary\\ NAS \end{tabular}          & \begin{tabular}[c]{@{}c@{}}NAS \cite{zoph2016neural} \\ MetaQNN \cite{baker2016designing} \\ EAS  \cite{zoph2016neural} \\  Block-QNN-S \cite{zhong2017practical} \end{tabular}                                                                                                                                                                                                       & \begin{tabular}[c]{@{}c@{}}93.91\\ 93.08\\ 95.77\\ 95.62\end{tabular}                                                                                & \begin{tabular}[c]{@{}c@{}}--\\ 77.14\\ --\\ 79.35\end{tabular}                                                                                 & \begin{tabular}[c]{@{}c@{}}22400\\ 100\\ 10\\ 90\end{tabular}                                               & \begin{tabular}[c]{@{}c@{}}2.02\\ 2.85\\ 0.16\\ 0.31\end{tabular}                                                                     & \begin{tabular}[c]{@{}c@{}}--\\ 4.22\\ --\\ 2.01\end{tabular}                                                                             \\ \hline
\multirow{2}{*}{ENAS}                                                   & \begin{tabular}[c]{@{}c@{}}Genetic-CNN \cite{xie2017genetic} \\ EVO-91a \cite{zhang2018finding} \\ EVO-91b \cite{zhang2018finding} \\ Large-scale evolution \cite{real2017large-scale} \\ Hierarchical evolution \cite{liu2017hierarchical}\\  AmoebaNet+cutout  \cite{real2019regularized} \\ CGP-CNN \cite{suganuma2017genetic} \\ CoDeepNEAT \cite{miikkulainen2019evolving} \\ CNN-GA \cite{sun2018automatically}\\ AE-CNN \cite{sun2018automatically} \\ AE-CNN+E2EPP \cite{sun2019surrogate} \end{tabular}                                                                          & \begin{tabular}[c]{@{}c@{}}94.61\\ 94.29\\ 94.81\\ 94.60\\ 96.37\\ 96.66\\ 94.02\\ 92.70\\ 95.22\\ 95.3\\ 94.70\end{tabular}                         & \begin{tabular}[c]{@{}c@{}}74.88\\ 71.3\\ 75.4\\ 77.00\\ --\\ --\\ --\\ --\\ 77.97\\ 77.6\\ 77.98\end{tabular}                                  & \begin{tabular}[c]{@{}c@{}}17\\ --\\ --\\ 2750\\ 300\\ 3150\\ 27\\ --\\ 37.5\\ 52\\ 8.5\end{tabular}         & \begin{tabular}[c]{@{}c@{}}1.32\\ 1.64\\ 1.12\\ 1.33\\ -0.44\\ -0.73\\ 1.91\\ 3.23\\ 0.71\\ 0.63\\ 1.23\end{tabular}                  & \begin{tabular}[c]{@{}c@{}}6.48\\ 10.06\\ 5.96\\ 4.36\\ --\\ --\\ --\\ --\\ 3.39\\ 3.76\\ 3.38\end{tabular}                               \\ \cline{2-7}
                                                                        & SI-ENAS                                                                                                                                                                                                                                                                          & 95.93                                                                                                                                                & 81.36                                                                                                                                           & 1.8                                                                                                         & --                                                                                                                                    & --                                                                                                                                        \\ \hline
\end{tabular}
\end{table*}

The experimental results in terms of the classification accuracy and consumed GPU days of all compared algorithms are presented in Table IV. In the table,  symbol “--” means that the corresponding result was not published. Note that all the results of the competitors in this table are extracted from the papers the methods were published.

From the results in Table IV, we can see that SI-ENAS is able to achieve better performance than all state-of-the-art manually designed DNNs. The performance enhancement of SI-ENAS on CIFAR100 is larger than 10\% compared to Maxout, VGG, Network in Network, Highways network, All-CNN and DSN, and larger than 5\% compared to DenseNet, ResNet(depth=101), and ResNet(depth=1202).

The classification accuracy of SI-ENAS is better than all four non-evolutionary NAS methods considered in this work, including NAS, MetaQNN, EAS, Block-QNN-S on both CIFAR10 and CIFAR100 datasets. Note also, NAS, MetaQNN, Block-QNN-S, and EAS have consumed 22400, 100, 90 and 10 GPU days, respectively, to achieve their best classification accuracies, while SI-ENAS has consumed only 1.8 GPU days.

Compared with the 11 ENAS methods, SI-ENAS performs better than Genetic-CNN, EVO, Large-scale evolution, CGP-CNN, CoDeepNEAT, CNN-GA, AE-CNN, AE-CNN+E2EPP, although slightly worse than Hierarchical evolution (0.44\%) and AmoebaNet+cutout (0.73\%) on CIFAR10. On CIFAR100, however, SI-ENAS has achieved the highest classification accuracy among all compared ENAS algorithms. Note again, Hierarchical evolution and AmoebaNet+cutout have consumed 300 and 3150 GPU days, respectively, while SI-ENAS has only consumed 1.8 GPU days.

\subsection{Effectiveness of the FR operation}


In the deep learning community, the channel attention mechanism is manually incorporated into a neural network. In this work, the channel attention module is part of the search operation in the encoded search space and will be adaptively incorporated into the network if it helps improve the learning performance. To check the effectiveness of the FR operation, we compare the neural network evolved by SI-ENAS with three other neural networks. One is a neural network evolved by SI-ENAS by switching off the FR operation, which is denoted by SI-ENAS without FR. The other two networks are obtained by mannually stacking the channel attention module, i.e., SE-block (SE) \cite{hu2018squeeze} and the residual channel attention module (RCAM) \cite{wang2017residual}, respectively, into SI-ENAS without FR as a connection structure between two blocks, which are denoted by SI-ENAS+SE and SI-ENAS+RCAM. For a fair comparison, all other settings are kept the same as those utilized for the experiment described in Subsection VI.A.

Table V presents the classification results of the four neural networks under comparison on CIFAR10 and CIFAR100. As we can see from Table V, SI-ENAS without FR, SI-ENAS+SE, and SI-ENAS+RCAM obtain the best classification accuracies of 93.72\%, 94.36\%, 94.85\% on CIFAR10, and 77.93\%, 79.13\%, 78.38\% on CIFAR100, respectively, while SI-ENAS with FR obtains the best classification accuracies of 95.93\% on CIFAR10 and 81.36\% on CIFAR100, respectively. We also note that SI-ENAS+RCAM achieves better classification performance than SI-ENAS+SE, indicating that the performance of the residual channel attention module is betters than the channel attention mechanism.

\begin{table}[htbp]
  \centering
  \caption{Classification accuracy (\%) of SI-ENAS with FR, SI-ENAS+SE, SI-ENAS+RCAM and SI-ENAS without FR on CIFAR10 and CIFAR100}
    \begin{tabular}{ccc}
    \toprule
    Algorithm & \multicolumn{1}{p{3.75em}}{CIFAT10} & \multicolumn{1}{p{4em}}{CIFAR100} \\
    \midrule
    SI-ENAS with FR & 95.93 & 81.36 \\
    SI-ENAS+SE & 94.36 & 78.38 \\
    SI-NAS+RCAM & 94.85 & 79.13 \\
    SI-ENAS without FR  & 93.72 & 77.93 \\
    \bottomrule
    \end{tabular}%
  \label{tab:addlabel}%
\end{table}%

\subsection{Effectiveness of Node inheritance}

\begin{figure}
\begin{minipage}[t]{0.5\linewidth}
\centering
\includegraphics[width=1.55in]{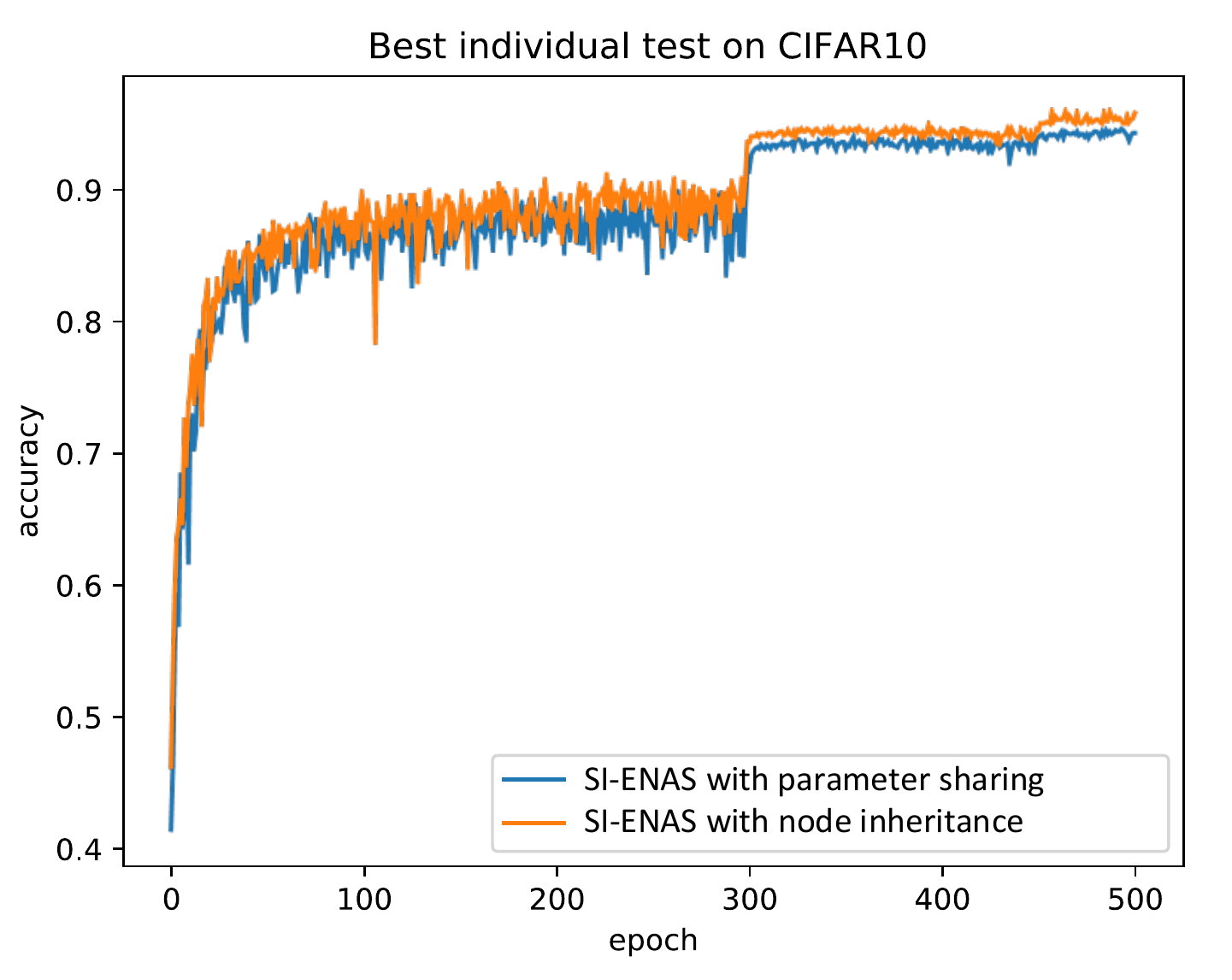}
\end{minipage}%
\begin{minipage}[t]{0.5\linewidth}
\centering
\includegraphics[width=1.56in]{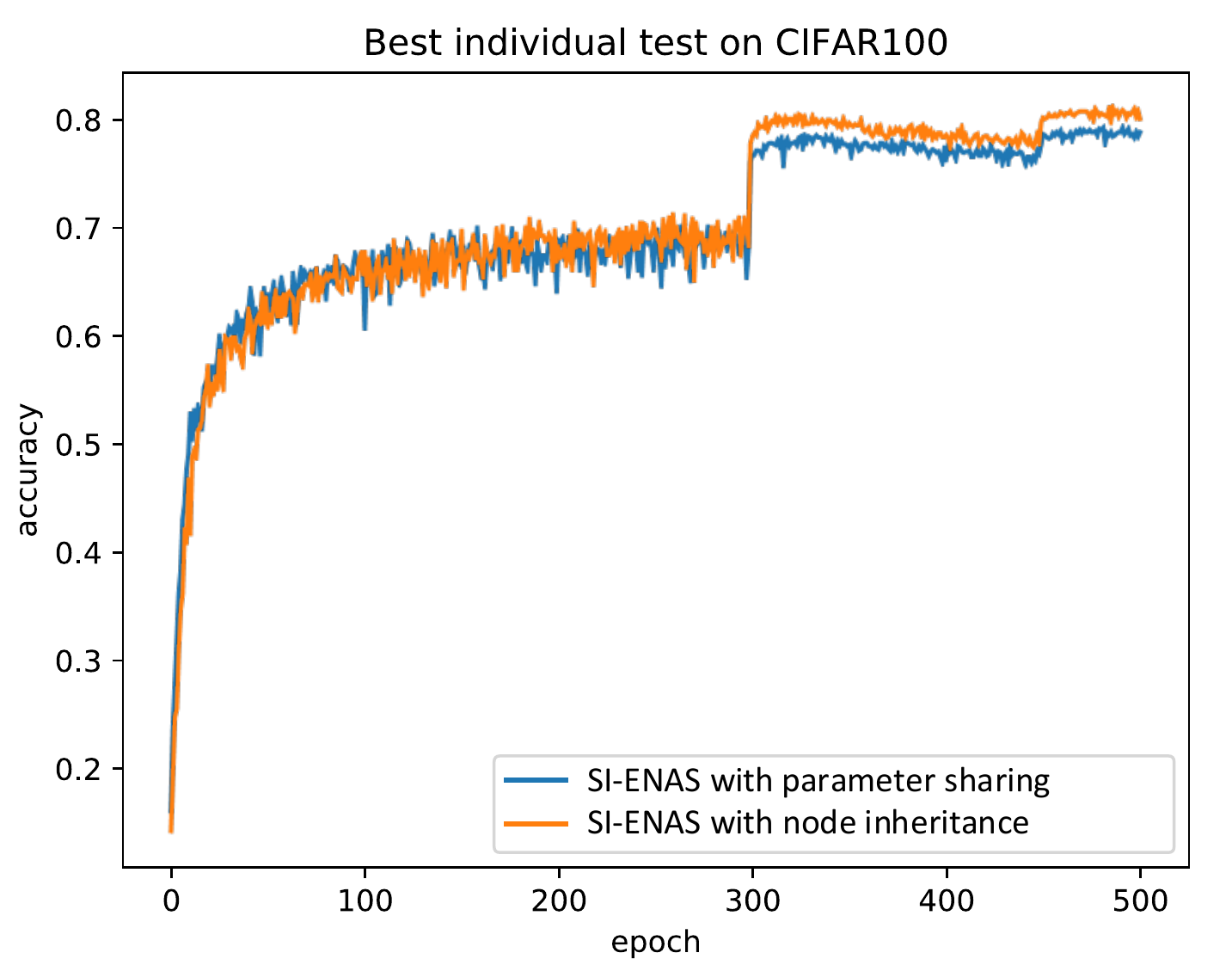}
\end{minipage}
\caption{Comparison of the validation accuracy of SI-ENAS with that of SI-ENAS with parameter sharing on CIFAR10 (left) and CIFAR100 (right).} 
\label{Fig_9}  
\end{figure}

To check the effectiveness of the proposed node inheritance mechanism, we compared it with the parameter sharing method \cite{pham2018efficient}, which was shown to speed up NAS by more than 1000 times. In other words, we replace the node inheritance strategy in SI-ENAS with the parameter sharing method during the evolution by forcing all offspring networks to copy weights from their parents. Recall that the parameter sharing strategy requires that each individual in the parent population has a separate set of weights. Therefore, once the offspring are generated by applying the genetic operations on the parents, all offspring models will share the weights of the best parent individual. This way, offspring models are never trained from scratch and the computation costs can also be reduced. For a fair comparison, all other settings are kept unchanged. The training processes are presented in Fig. \ref{Fig_9} and the final accuracies are listed in Table VI.

\begin{table}[htbp]
  \centering
  \caption{Classification accuracy (\%) of SI-ENAS with node inheritance and SI-ENAS with parameter sharing on the CIFAR10 and CIFAR100 datasets.}
    \begin{tabular}{ccc}
    \toprule
    Algorithm & \multicolumn{1}{p{4.415em}}{CIFAR10} & \multicolumn{1}{p{4.04em}}{CIFAR100} \\
    \midrule
    SI-ENAS with node inheritance & 95.93 & 81.36 \\
    SI-ENAS with parameter sharing  & 94.82 & 79.59 \\
    \bottomrule
    \end{tabular}%
  \label{tab:addlabel}%
\end{table}%

As shown in Table VI, SI-ENAS with node inheritance achieves a classification accuracy of 95.93\% and 81.36\%, respectively, on CIFAR 10 and CIFAR100. By contrast, SI-ENAS with parameter sharing achieves a classification accuracy of 94.82\% and 79.59\%, respectively. This is due to the fact that parameter sharing forces all offspring models to share a set of weights from the best parent individual, and consequently the offspring models do not inherit the parameters from their own parents. As a result, the estimated fitness value of the offspring networks is subject to big errors, which may prevent the EA from finding the best neural architecture.

\subsection{Neural architecture transferring}

Here, we examine if the network architecture optimized on CIFAR10, a relatively small dataset, can be directly used to effectively learn bigger datasets such as CIFAR100 and SVHN.

To observe the change of the transferring performance of the neural networks during the optimization, we divide the evolutionary process into three stages based on the changes of the learning rate. Specifically, the initial population is denoted as stage 0, generations 1 to 149 is denoted as stage 1, generations 150 to 224 is denoted as stage 2, generations 225 to 299 as stage 3. Then, we evaluate the performance of the best individual at each stage on all data in the SVHN and CIFAR100 datasets, respectively. Note that an individual is randomly picked from the initial population at stage 0.


\begin{table}[htbp]
  \centering
  \caption{The performance of the neural architecture optimized on CIFAR10 and tested on CIFAR100 and SVHN, respectively, in various search stages. CIFAR100-CIFAR100 denotes the network optimized on the training and validation sets of CIFAR100 and evaluated on the test dataset of CFAR100, and SVHN-SVHN the network optimized on the training and validation datasets of SVHN, and tested on the validation data of SVHN.}
\begin{tabular}{c|c|c|c}
\hline
Test dataset              & Search dataset & Individual                                                                    & Accuracy                                                              \\ \hline
\multirow{2}{*}{SVHN}     & CIFAR10        & \begin{tabular}[c]{@{}c@{}}Stage 0\\ Stage 1\\ Stage 2\\ Stage 3\end{tabular} & \begin{tabular}[c]{@{}c@{}}97.16\\ 97.74\\ 98.02\\ 98.31\end{tabular} \\ \cline{2-4}
                          & SVHN           & SVHN-SVHN                                                                     & 98.43                                                                 \\ \hline
\multirow{2}{*}{CIFAR100} & CIFAR10        & \begin{tabular}[c]{@{}c@{}}Stage 0\\ Stage 1\\ Stage 2\\ Stage 3\end{tabular} & \begin{tabular}[c]{@{}c@{}}74.83\\ 75.96\\ 79.84\\ 80.13\end{tabular} \\ \cline{2-4}
                          & CIFAR100       & CIFAR100-CIFAR100                                                             & 81.36                                                                 \\ \hline
\end{tabular}
\end{table}

The experimental results are presented in Table VII. From the table, we can see that the best individual trained on CIFAR10 is able to achieve a classification accuracy of 98.31\% and 80.13\% on CIFAR100 and SVHN, respectively, which is slightly lower than the network trained on CIFAR100 (98.43\%) and SVHN (81.36\%). From these results, we can conclude that the neural network structure optimized by the proposed SI-ENAS has a promising capability to be transferred to different datasets.

\section{Conclusion and Future Work}

This work proposes a fast ENAS framework that is well suited for implementation on devices with limited computation resources. The computation costs of fitness evaluations is dramatically reduced by two related strategies, sampled training of the parent individuals and node inheritance of the offspring individuals. To further improve the expression ability in evolving large neural networks, the multi-scale feature reconstruction convolutional operation is encoded into search space. Our experimental results demonstrate that the SI-ENAS can effectively speed up the evolutionary architecture search and achieve very promising classification accuracy. Finally, we transfer the neural architecture optimized on the smaller dataset CIFAR10 to larger CIFAR100 and SVHN datasets and encouraging experimental results are obtained.

Although the SI-ENAS is competitive to design a high-performance neural network architectures, its search capability remains to be examined on larger datasets and real-world problems. We will also extend the proposed method to NAS for optimizing properties  in addition to classification accuracy, such as robustness and explanability. In this case, multi-objective evolutionary neural architecture search may play an important role.

\bibliographystyle{IEEEtran}

\bibliography{cas-refs}

\end{document}